%% file: main.tex
\documentclass{article}
\usepackage[T1]{fontenc}
\usepackage{amsmath,graphicx,mlspconf}
\usepackage{amsthm}
\usepackage{subcaption}
\usepackage{booktabs}
\usepackage{nicefrac}
\usepackage{multirow}
\usepackage{amsfonts}
\usepackage{lipsum}
\usepackage{float}
\usepackage{xcolor}

\usepackage[linesnumbered,ruled,vlined]{algorithm2e}

\SetKwComment{Comment}{\color{green!50!black}// }{}

\SetKwProg{Function}{function}{}{}

\usepackage[acronym]{glossaries}
\usepackage{wrapfig}
\usepackage{nicefrac}
\usepackage{fontawesome}
\graphicspath{{./Figures}}
\usepackage[hidelinks]{hyperref}
\usepackage{comment}

\newcommand{\arginf}[1]{\underset{#1}{\text{arginf}}\,}
\newcommand{\argmin}[1]{\underset{#1}{\text{argmin}}\,}
\newcommand{\argmax}[1]{\underset{#1}{\text{argmax}}\,}
\newcommand{\expectation}[1]{\underset{#1}{\mathbb{E}}}

\newcommand{\iid}[0]{\overset{\text{i.i.d.}}{\sim}}

\input{glossaries.tex}

\title{Online Multi-Source Domain Adaptation through Gaussian Mixtures and Dataset Dictionary Learning}
\name{Eduardo Fernandes Montesuma$^{1}$, Stevan Le Stanc$^{1,2}$, Fred Ngol\'e Mboula$^{1}$}
\address{$^{1}$Université Paris-Saclay, CEA, List, F-91120 Palaiseau, France\\
$^{2}$Ecole Centrale de Lyon, 69134 Ecully, France}

\begin{document}

\maketitle

\begin{abstract}
    This paper addresses the challenge of online multi-source domain adaptation (MSDA) in transfer learning, a scenario where one needs to adapt multiple, heterogeneous source domains towards a target domain that comes in a stream. We introduce a novel approach for the online fit of a Gaussian Mixture Model (GMM), based on the Wasserstein geometry of Gaussian measures. We build upon this method and recent developments in dataset dictionary learning for proposing a novel strategy in online MSDA. Experiments on the challenging Tennessee Eastman Process benchmark demonstrate that our approach is able to adapt \emph{on the fly} to the stream of target domain data. Furthermore, our online GMM serves as a memory, representing the whole stream of data\footnote{This is a pre-print. This paper was accepted at the IEEE International Workshop on Machine Learning for Signal Processing 2024}.
\end{abstract}
\section{Introduction}\label{sec:intro}

Modern machine learning systems rely on rich, large-scale datasets~\cite{zhou2023deep}. At the same time, these systems are often confronted with problem of \emph{distributional shift}~\cite{quinonero2008dataset}, a problem in which test data comes from a different, but related probability distribution. In a realistic scenario, target domain data is available in a stream, rather than in a stored and annotated fashion. This motivates the field of \emph{online \gls{da}}. In parallel, \emph{multi-source \gls{da}} considers that the source domain data actually come from multiple, heterogeneous domains.

An example of application of incremental \gls{da} is automatic fault diagnosis~\cite{montesuma2022cross}. In this problem, one wants to determine, from sensor readings, whether a system is in its nominal state, or in some type of faulty state. Nonetheless, in order to reliably collect a large dataset of faults, the system needs to fail many times, which may pose security hazards and economic losses. A possible solution is to rely on historical data or simulations, but this leads to a distributional shift between the existing data, and the data that the system needs to predict on the fly. Hence, incremental \gls{da} is a good candidate to enhance the performance of automatic fault diagnosis systems.

In the context of \gls{da}, a prominent framework is \gls{ot}~\cite{peyre2019computational,montesuma2023recent}, which is a mathematical theory concerned with the displacement of mass at least effort. In this paper, we are particularly interested in the \gls{dadil} framework proposed by~\cite{montesuma2023learning}, especially its \gls{gmm} formulation~\cite{montesuma2024faster}, which learns to interpolate probability measures in a Wasserstein space through dictionary learning. So far, these methods assume that all data is available during the training process. In this work we take a step forward, and consider the adaptation towards the target domain in an \emph{online} fashion.

Our contributions are twofold. First, we propose a novel strategy for the online learning of \glspl{gmm}, based on the Wasserstein geometry over the space of Gaussian measures. Second, we show that through our online \gls{gmm} algorithm, we can reliably learn a dictionary of \glspl{gmm} for online \gls{msda}. While previous works have considered the slightly related field of class incremental fault diagnosis~\cite{xie2012dynamic,guan2022model,fu2022broad}, ours is the first to consider online cross-domain fault diagnosis, especially through \gls{msda}.

This paper is organized as follows. Section~\ref{sec:background} covers the background to our proposed methods, namely Gaussian mixtures, optimal transport and domain adaptation. Section~\ref{sec:methodology} covers our proposed incremental \gls{gmm}, and incremental dictionary learning methods. Section~\ref{sec:experiments} covers our experiments on a toy dataset, and cross-domain fault diagnosis on the \gls{tep} benchmark~\cite{reinartz2021extended,montesuma2023multi}. Finally, section~\ref{sec:conclusion} concludes this paper.

\section{Background}\label{sec:background}

\subsection{Gaussian Mixture Models}

\glspl{gmm} are a type of probabilistic model that can handle data with sub-populations. Let $\mathcal{N}(\mu, \Sigma)$ denote a Gaussian measure over $\mathbb{P}(\mathbb{R}^{d})$, a \gls{gmm} with $K \in \mathbb{N}$ components is,
\begin{align*}
    P_{\theta}(\mathbf{x}) = \sum_{k=1}^{K}\pi_{k}^{(P)}P_{k}\text{, and }P_{k}=\mathcal{N}(\mu_{k}^{(P)},\Sigma_{k}^{(P)}),
\end{align*}
where $\theta = \{(\pi_{k}^{(P)},\mu_{k}^{(P)},\Sigma_{k}^{(P)})\}_{k=1}^{K}$. For a general measure $P \in \mathbb{P}(\mathbb{R}^{d})$, a \gls{gmm} can be fit to data $\{\mathbf{x}_{i}^{(P)}\}_{i=1}^{n}$ via maximum-likelihood estimation,
\begin{align}
    \theta^{\star} = \argmax{\theta}L(\theta):=\dfrac{1}{n}\sum_{i=1}^{n}\log P_{\theta}(\mathbf{x}_{i}^{(P)}).\label{eq:maximum_likelihood}
\end{align}
While equation~\ref{eq:maximum_likelihood} has no closed-form solution, this problem can be solved via the celebrated \gls{em} algorithm~\cite{dempster1977maximum}. We denote by $\text{EM}(\mathbf{X}^{(P)},K)$, the operation of fitting a $K-$component \gls{gmm} via the \gls{em} algorithm on data $\mathbf{X}^{(P)} \in \mathbb{R}^{n \times d}$.

\vspace{1mm}

\noindent\textbf{Model selection.} An important hyperparameter in \glspl{gmm} is the number of components $K$, which controls the complexity of the model. This parameter may be determined via the \gls{bic}~\cite{schwarz1978estimating},
\begin{align}
    \text{BIC}(P_{\theta}) = |P_{\theta}|\log(n) - 2\log(L(\theta)),\label{eq:bic}
\end{align}
where $|P_{\theta}|$ denotes the number of parameters in the \gls{gmm}, $n$ denotes the number of data points and $L(\theta)$ denotes the likelihood of the learned \gls{gmm}.

\vspace{1mm}

\noindent\textbf{Labeled \glspl{gmm}.} As in~\cite{montesuma2024faster}, we consider labeled \glspl{gmm}, i.e., to each $P_{k}$ there is an associated label $\tilde{\mathbf{y}}_{k}^{(P)}$. These labels are defined by fitting a \gls{gmm} on the conditional measure $P_{y} = P(X|Y=y)$. This is done by performing an \gls{em} on the data $\{\mathbf{x}_{i}^{(P)}\}_{i:y_{i}^{(P)}}=y$. Based on the \gls{gmm}, we can classify samples using \gls{map} estimation,
\begin{align}
    \hat{y} = \argmax{j=1,\cdots,n_{c}}\sum_{k=1}^{K}P_{\theta}(y=j|k)P_{\theta}(k|\mathbf{x}),\label{eq:map}
\end{align}
where $P_{\theta}(y=j|k) = \tilde{y}_{kj}^{(P)}$.

\subsection{Optimal Transport}\label{sec:ot}

In this section we give a brief introduction to \gls{ot}. We refer readers to~\cite{peyre2019computational,montesuma2023recent} for recent and comprehensive introductions to the topic. \gls{ot} is a mathematical theory concerned with the transportation of mass at least effort. In its modern formulation, it describes the transportation between probability measures. For a set $\mathcal{X}$ (e.g., $\mathbb{R}^{d}$), let $P$ and $Q$ be measures in $\mathbb{P}(\mathcal{X})$. A transport plan is a measure $\gamma \in \mathbb{P}(\mathcal{X}^{2})$ that preserves mass, i.e.,
\begin{align*}
    \int_{\mathcal{X}}\gamma(x,B)dx = Q(B)\text{, and, }\int_{\mathcal{X}}\gamma(A,x)dx = P(A),
\end{align*}
or $\gamma \in \Gamma(P, Q)$, in short. Let $c:\mathcal{X}\times\mathcal{X}\rightarrow\mathbb{R}$ be a ground-cost. The \gls{ot} problem is given by,
\begin{align}
    \gamma^{\star} = \arginf{\gamma \in \Gamma(P, Q)}\int_{\mathcal{X}}\int_{\mathcal{X}}c(x_{1},x_{2})d\gamma(x_{1},x_{2}),\label{eq:kantorovich_ot}
\end{align}
which is a linear program with respect the transport plan $\gamma$. Let $\alpha \in [1,\infty)$, and $(\mathcal{X},d)$ be a metric space. When $c(x_{1},x_{2}) = d(x_{1},x_{2})^{\alpha}$, one may define a distance associated with \gls{ot} in equation~(\ref{eq:kantorovich_ot}),
\begin{align}
    \mathcal{W}_{\alpha}(P, Q)^{\alpha} = \int_{\mathcal{X}}\int_{\mathcal{X}}d(x_{1},x_{2})^{\alpha}d\gamma^{\star}(x_{1},x_{2}),
\end{align}
called \emph{Wasserstein distance}. This distance lifts the metric $d$ on $\mathcal{X}$ to a metric $\mathcal{W}_{\alpha}$ on $\mathbb{P}(\mathcal{X})$. Henceforth we assume $\mathcal{X} = \mathbb{R}^{d}$, $\alpha=2$ and $d(\mathbf{x}_{1},\mathbf{x}_{2}) = \lVert \mathbf{x}_{1} - \mathbf{x}_{2} \rVert_{2}$. Next, we describe two particular cases in which \gls{ot} is tractable.

\vspace{1mm}

\noindent\textbf{Gaussian OT.} When $P$ and $Q$ are Gaussian measures, i.e., $P=\mathcal{N}(\mu^{(P)},\Sigma^{(P)})$ and $Q=\mathcal{N}(\mu^{(Q)},\Sigma^{(Q)})$, \gls{ot} has a closed-form solution~\cite{takatsu2011wasserstein}. In this case the Wasserstein distance takes the form,
\begin{align*}
    \mathcal{W}_{2}(P, Q)^{2} &= \lVert \mu^{(P)} - \mu^{(Q)} \rVert^{2}_{2} + \mathcal{BU}(\Sigma^{(P)},\Sigma^{(Q)}),
\end{align*}
where $\mathcal{BU}(\mathbf{A}, \mathbf{B})$ denotes the Bures metric between covariance matrices~\cite{takatsu2011wasserstein}. If the $P$ and $Q$ are axis-aligned, i.e., $\Sigma^{(P)}$ and $\Sigma^{(Q)}$ are diagonal matrices with standar deviation vectors $\sigma^{(P)}, \sigma^{(Q)} \in \mathbb{R}^{d}_{+}$,
\begin{align}
    \mathcal{W}_{2}(P, Q)^{2} &= \lVert \mu^{(P)} - \mu^{(Q)} \rVert^{2}_{2} + \lVert \sigma^{(P)} - \sigma^{(Q)} \rVert^{2}_{2},\label{eq:gauss_w2}
\end{align}
that is, with a Wasserstein metric, the space of Gaussian measures parametrized by $(\mu, \sigma)$ is isomorphic to $\mathbb{R}^{2d}$.

\vspace{1mm}

\noindent\textbf{GMM-OT.} As studied by~\cite{delon2020wasserstein}, \gls{ot} between two \glspl{gmm} $P$ and $Q$ is tractable, when $\gamma$ is further restricted to be a \gls{gmm}. Let $\omega \in \mathbb{R}^{K_{P} \times K_{Q}}$ be an \gls{ot} plan between \emph{components}, then,
\begin{align}
    \omega^{\star} &= \argmin{\omega \in \Gamma(\pi^{(P)}, \pi^{(Q)})}\sum_{k_{1}=1}^{K_{P}}\sum_{k_{2}=1}^{K_{Q}}\omega_{k_{1},k_{2}}\mathcal{W}_{2}(P_{k_{1}},Q_{k_{2}})^{2},
\end{align}
where $P_{k_{1}}$ with weight $\pi_{k_{1}}^{(P)}$ denote the $k_{1}-$th component of the \gls{gmm} $P$ (resp. $k_{2}$ and $Q$). As shown in~\cite{delon2020wasserstein}, there is an \gls{ot} plan \emph{between samples}, $\gamma^{\star}$, associated with $\omega^{\star}$. This formulation is used to define a Wasserstein-like distance,
\begin{align}
    \mathcal{MW}_{2}(P, Q) &= \sum_{k_{1}=1}^{K_{P}}\sum_{k_{2}=1}^{K_{Q}}\omega_{k_{1},k_{2}}^{\star}\mathcal{W}_{2}(P_{k_{1}},Q_{k_{2}})^{2},\label{eq:mw2}
\end{align}
which is an hierarchical \gls{ot} distance, i.e., it depends on an inner transportation problem between Gaussian measures.

\vspace{1mm}

\noindent\textbf{Barycenters.} Given measures $\mathcal{P} = \{P_{1},\cdots,P_{C}\}$ in $\mathbb{P}(\mathbb{R}^{d})$, we can define the mixture-Wasserstein barycenter as~\cite{agueh2011barycenters,delon2020wasserstein},
\begin{align}
    \mathcal{B}(\lambda,\mathcal{P}) = \argmin{B \in \mathbb{P}(\mathbb{R}^{d})}\sum_{c=1}^{C}\lambda_{c}\mathcal{MW}_{2}(B,P_{c})^{2}.
\end{align}
When the \glspl{gmm} have labels, i.e., to each component $P_{k_{1}}$ there is an associated label $\tilde{\mathbf{y}}_{k_{1}}^{(P)}$, we add a term, $\beta\lVert \mathbf{y}_{k_{1}}^{(P)} - \mathbf{y}_{k_{2}}^{(Q)} \rVert_{2}^{2}$, to the ground cost, for $\beta > 0$. This latter parameter controls the relevance of labels for the ground-cost. We denote the associated distance $\mathcal{SMW}_{2,\beta}$, for supervised mixture-Wasserstein distance~\cite[Sec. 3.2.]{montesuma2024faster}.

\subsection{Learning Theory and Domain Adaptation}

In this paper, we consider the domain adaptation for classification, under the \gls{erm} framework of~\cite{vapnik2013nature}. For a probability measure $P$, a loss $\mathcal{L}$ and a ground-truth labeling function $h_{0}$, the risk of a classifier $h \in \mathcal{H}$ is given by,
\begin{align}
    \mathcal{R}_{P}(h) &= \expectation{\mathbf{x} \sim P}[\mathcal{L}(h(\mathbf{x}), h_{0}(\mathbf{x}))].\label{eq:risk}
\end{align}
The risk minimization strategy consists of finding $h^{\star} \in \mathcal{H}$ that minimizes $\mathcal{R}_{P}(h)$. However, this requires knowing $P$ and $h_{0}$, which is not feasible in most cases. As a result, one may approximate equation~(\ref{eq:risk}) empirically,
\begin{align}
    \hat{\mathcal{R}}_{P}(h) &= \dfrac{1}{n}\sum_{i=1}^{n}\mathcal{L}(h(\mathbf{x}_{i}^{(P)}), \mathbf{y}_{i}^{(P)}),\label{eq:empirical_risk}
\end{align}
where $\mathbf{x}_{i}^{(P)} \iid P$ and $\mathbf{y}_{i}^{(P)} = h_{0}(\mathbf{x}_{i}^{(P)})$. In the context of equation~(\ref{eq:empirical_risk}), \gls{erm} consists of minimizing $\hat{\mathcal{R}}_{P}$ with respect $h \in \mathcal{H}$.

Under the assumption that new data points are sampled from $P$, it is possible to bound the true risk $\mathcal{R}_{P}$ with respect the empirical risk $\hat{\mathcal{R}}_{P}$, which poses the theoretical grounds for generalization. However, in many practical cases, we want to apply a classifier on data with slightly different properties. For instance in object recognition, one may collect images from the web to consitute a large labeled dataset of objects, which constitutes the source domain. For the target domain, consider photos taken by a phone, which have slightly different properties than the ones in the source domain (e.g., illumination, background), but still represent the same objects.

\section{Proposed Method}\label{sec:methodology}

\noindent\textbf{Problem Statement.} In this paper, we consider the problem of online \gls{msda}. In classic \gls{msda}, one has a set of $N_{S}$, labeled probability measures, i.e., $\mathcal{Q}_{S} = \{\hat{Q}_{S_{\ell}}\}_{\ell=1}^{N_{S}}$. The challenge is to learn a classifier on an unlabeled target measure $\hat{Q}_{T}$, using labeled samples from the measures in $\mathcal{Q}_{S}$, and unlabeled samples on $\hat{Q}_{T}$. In this paper, we seek to do so in an \emph{on-line} fashion. In this setting, we assume that the source domain samples are available offline, but the samples from the target measure arrive in a stream, and are seen only once. We propose to tackle this problem by representing the target domain measure via a \gls{gmm}. Next, we divide our discussion into the two components of our method: online \gls{gmm} and online \gls{gmm}-\gls{dadil}.

\subsection{Online Gaussian Mixture Modeling}\label{sec:ogmm}

We propose a novel method for the online learning of \glspl{gmm}, based on~\cite{acevedo2017multivariate}. Our work is different from theirs, as we use the Wasserstein geometry over the Gaussian family (described in section~\ref{sec:ot}). Note that data comes in a stream.

In a nutshell, our strategy consists of progressively growing the mixture model. We start with $K_{min}$ components, fit to the first batch of data. We then grow the \gls{gmm}, by fitting a mixture model on the new batch of data (\texttt{\bfseries get\_best\_gmm}), then appending the new components to the existing ones. Eventually, this process creates more components than the maximum allowed, i.e., $K \geq K_{max}$. We then \emph{compress} the \gls{gmm} (\texttt{\bfseries compress\_gmm}) by finding the most similar components, then merging them (\texttt{\bfseries gauss\_merge}). Our overall algorithm is shown in algorithm~\ref{alg:online_gmm_fit}. We now discuss each component of our strategy.

\input{algorithms/online_gmm_fit}

\noindent\textbf{Fitting GMM to each batch.} For a new batch of data, we need to find a \gls{gmm} to represent it. We show the strategy in Algorithm~\ref{alg:get_best_gmm}. This process is challenging because we need to fit a \gls{gmm} with only a few data points. Besides assuming axis-aligned \glspl{gmm}, we limit the number of components fit to the data to the set $\{k_{1},\cdots,k_{2}\}$. As shown in Algorithm~\ref{alg:online_gmm_fit}, line 4, we use $k_{1} := 1$ and $k_{2} := \Delta K$, where $\Delta K$ is the maximum number of components fit to each batch. Note that we need $\Delta K < |\mathbf{X}^{(P)}_{t}|$. To determine the best possible fit to the batch, we use the \gls{bic} score in equation~(\ref{eq:bic}).

\input{algorithms/get_best_gmm}

\begin{figure}[ht]
    \centering
    \begin{subfigure}{0.32\linewidth}
        \includegraphics[width=\linewidth]{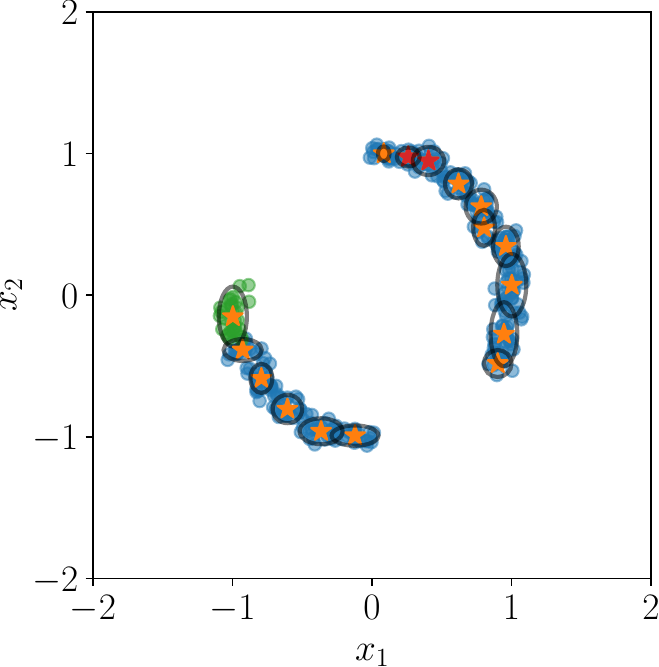}
        \caption{}
    \end{subfigure}\hfill
    \begin{subfigure}{0.32\linewidth}
        \includegraphics[width=\linewidth]{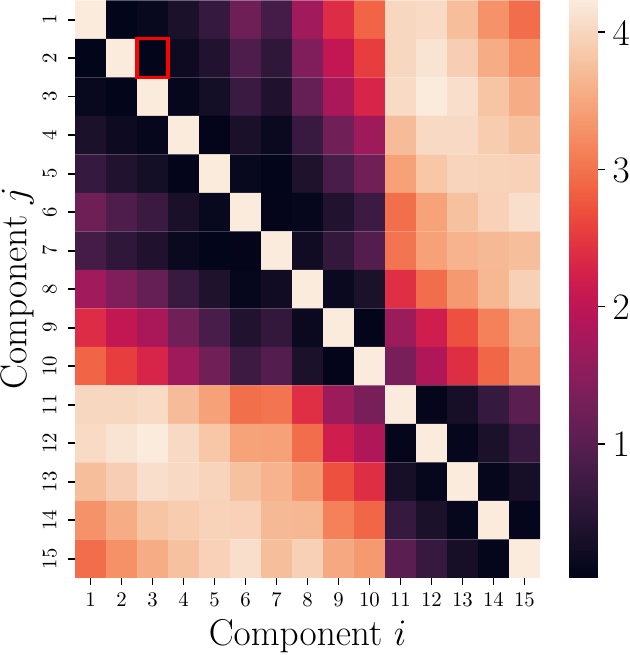}
        \caption{}
    \end{subfigure}\hfill
    \begin{subfigure}{0.32\linewidth}
        \includegraphics[width=\linewidth]{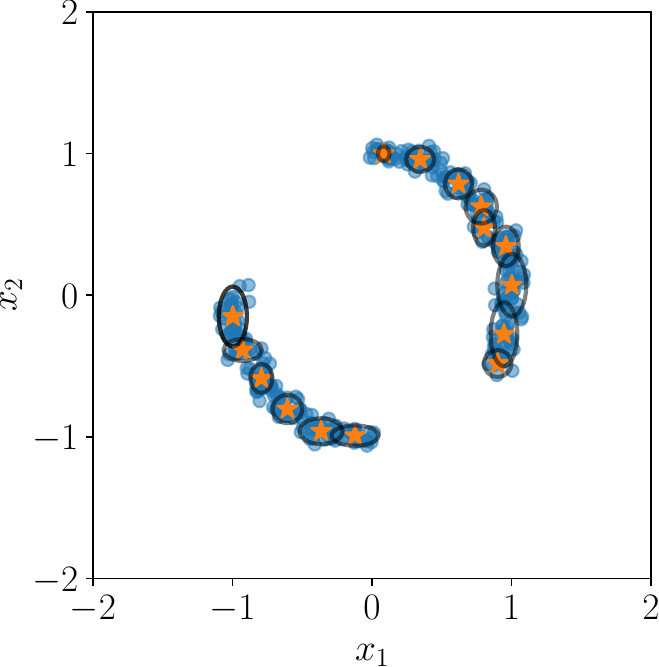}
        \caption{}
    \end{subfigure}
    \caption{Illustration of the proposed compression mechanism. In (a), a novel batch of data arives, making $K > K_{max}$. As a result, we compute the pairwise Wasserstein distance between \gls{gmm} components (b). We then take the two closest components (a, in red), and merge them. In (c), we show in red the resulting component of the merging process.}
    \label{fig:gmm_merge}
\end{figure}

\noindent\textbf{Compressing the GMM.} Through lines 4 and 5 in algorithm~\ref{alg:online_gmm_fit}, it is possible that the \gls{gmm} grows beyond $K_{max}$. As a result, we need to reduce the number of components within the \gls{gmm}. The process of \gls{gmm} compression consists of two components. First, we calculate the distances $W_{ij} = \mathcal{W}_{2}(P_{i}, P_{j})$ between pairs of components, while setting $W_{ii} = +\infty$. We do so to avoid merging a component with itself. We then select the components $i^{\star}$ and $j^{\star}$ that are the most similar and merge them using Algorithm~\ref{alg:merge_gaussians}. We repeat this process until the number of components $|P|$ reaches $K_{max}$. The compressing mechanism is shown in Fig.~\ref{fig:gmm_merge}.

\input{algorithms/compress_gmm}

\noindent\textbf{Merging Gaussians.} Given indices $(i, j)$, we need to combine components $P_{i}$ and $P_{j}$ with weights $\pi_{i}^{(P)}$ and $\pi_{j}^{(P)}$. To preserve the total mass of the \gls{gmm} (i.e., $\sum \pi_{i}^{(P)} = 1$), we set the new weight $\pi = \pi_{i}^{(P)} + \pi_{j}^{(P)}$. Next, we need to combine the mean and standard deviation parameters. We do so through Wasserstein barycenters~\cite{agueh2011barycenters}, i.e., $(\mu, \sigma) = \mathcal{B}(\lambda, \{P_{i}, P_{j}\})$. The barycentric coordinates vector $\lambda = (\lambda_{1}, \lambda_{2})$ corresponds to the relative weights of components $P_{i}$ and $P_{j}$, that is,
\begin{align*}
    \lambda_{1} = \frac{\pi_{i}^{(P)}}{\pi_{i}^{(P)}+\pi_{j}^{(P)}}\text{, and }\lambda_{2} = \frac{\pi_{j}^{(P)}}{\pi_{i}^{(P)}+\pi_{j}^{(P)}}.
\end{align*}
Since \gls{ot} is associated with an Euclidean geometry~\cite{peyre2019computational,montesuma2023recent} over the space $(\mu, \sigma) \in \mathbb{R}^{2d}$, the resulting parameters $\mu$ and $\sigma$ correspond to a simple weighting with $\lambda_{1}$ and $\lambda_{2}$, as shown in lines 4 and 5 of Algorithm~\ref{alg:merge_gaussians}.

\input{algorithms/merge_gaussians}

As a result, our method is quite different from the one proposed in~\cite{acevedo2017multivariate}. First, our merging process (Algorithm~\ref{alg:compress_gmm}) considers all components in the \gls{gmm}, whereas~\cite{acevedo2017multivariate} only removes components from the new \gls{gmm}. Second, we use the Wasserstein distance for comparing and merging components (algorithms~\ref{alg:compress_gmm} and~\ref{alg:merge_gaussians}), while~\cite{acevedo2017multivariate} uses the Kullback-Leibler divergence. While the Wasserstein distance is associated with an Euclidean geometry over $(\mu,\sigma) \in \mathbb{R}^{2d}$, the Kullback-Leibler is associated with an hyperbolic geometry~\cite[Chapter Remark 8.2]{peyre2019computational}.

\subsection{Online GMM-DaDiL}

In this section we introduce our novel online \gls{msda} strategy, which is based on our online \gls{gmm} algorithm and the \gls{gmm}-\gls{dadil} framework of~\cite{montesuma2024faster}. These authors introduced the notion of a dictionary $(\Lambda,\mathcal{P})$ of barycentric coordinate vectors $\Lambda = (\lambda_{1},\cdots,\lambda_{N_{S}},\lambda_{T})$, and atoms $\mathcal{P} = \{P_{1}, \cdots, P_{C}\}$. The atoms are \glspl{gmm}, i.e., $P_{c} = \sum_{k=1}^{K}\pi^{(P_{c})}_{k}P_{c,k}$ with learnable parameters $\mu_{k}^{(P_{c})}$ and $\sigma_{k}^{(P_{c})}$. The goal of \gls{gmm}-\gls{dadil} is expressing each $Q_{\ell}$ as a mixture-Wasserstein barycenter of $\mathcal{P}$, weighted by the barycentric coordinates $\lambda_{\ell}$~\cite[Algorithm 2]{montesuma2024faster}.

\vspace{1mm}

\noindent\textbf{Online GMM-DaDiL.} Given the offline data from source domains, we learn a \gls{gmm} on each source, denoted by $Q_{S_{\ell}}$. We then use the \gls{gmm} as a replay memory~\cite{zhou2023deep} during the online learning process. As we do not have access to the complete target domain data, we learn a \gls{gmm} on this domain through our \gls{ogmm} strategy (Algorithm~\ref{alg:online_gmm_fit}). At time step $t$, we denote this \gls{gmm} as $Q_{T}^{(t)}$. After each update on this measure, we update the \gls{gmm} using,
\begin{equation}
\begin{aligned}
    \mathcal{L}(\Lambda,\mathcal{P}) = &\mathcal{MW}_{2}(Q_{T}^{(t)}, \mathcal{B}(\lambda_{T}, \mathcal{P}))^{2} +\\ &\sum_{\ell=1}^{N_{S}}\mathcal{SMW}_{2}(Q_{S_{\ell}}, \mathcal{B}(\lambda_{\ell}, \mathcal{P}))^{2}.
\end{aligned}\label{eq:o-gmm-dadil}
\end{equation}
An interesting advantage of this strategy is that optimization can be carried out even after the data stream has ended, as the target \gls{gmm} serves as a representation for the whole history of received batches.

\section{Experiments}\label{sec:experiments}

\subsection{Online Gaussian Mixture Modeling}

\begin{figure}[ht]
    \centering
    \begin{subfigure}{0.32\linewidth}
        \includegraphics[width=\linewidth]{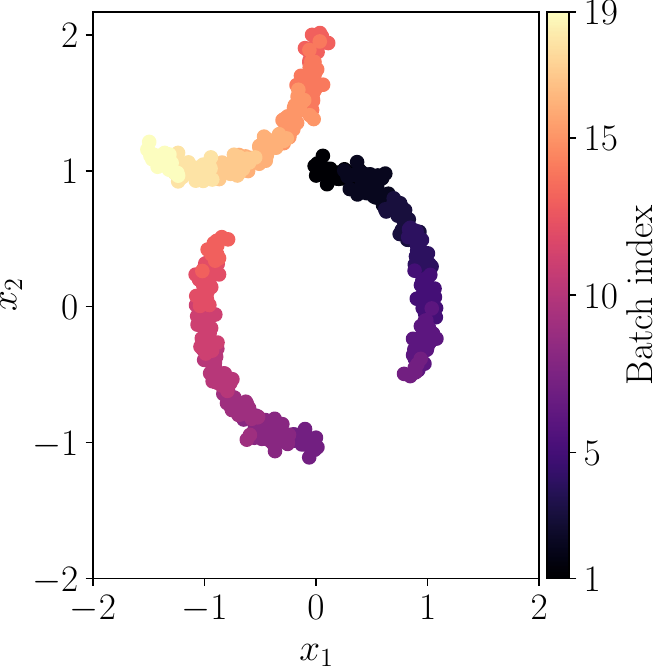}
        \caption{Order of batches.}
    \end{subfigure}\hfill
    \begin{subfigure}{0.32\linewidth}
        \includegraphics[width=\linewidth]{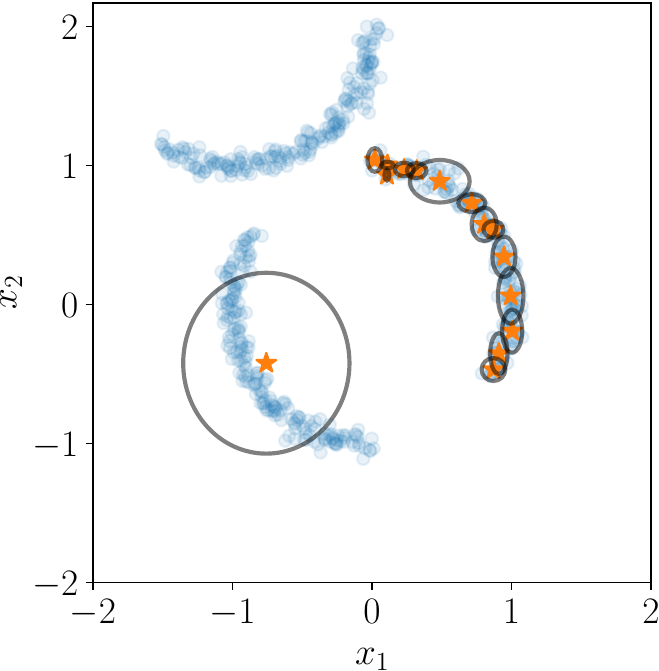}
        \caption{$it=10$ ($KL$).}
    \end{subfigure}\hfill
    \begin{subfigure}{0.32\linewidth}
        \includegraphics[width=\linewidth]{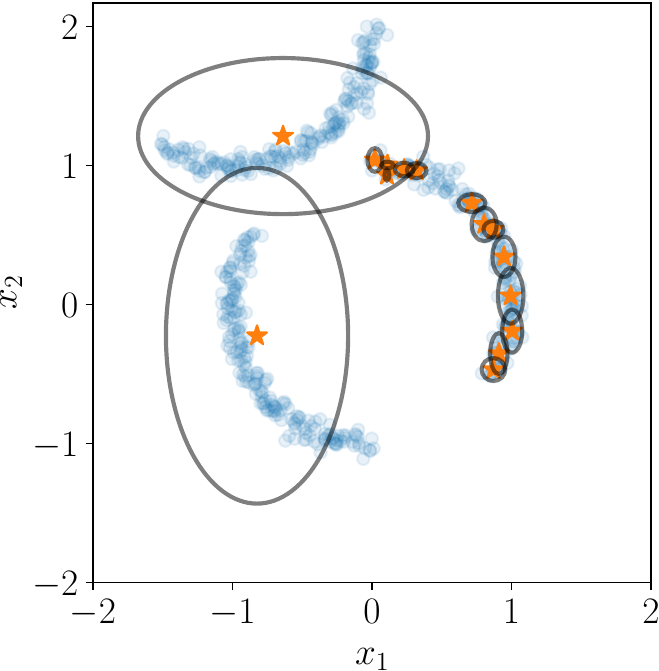}
        \caption{$it=18$ ($KL$).}
    \end{subfigure}\\
    \begin{subfigure}{0.32\linewidth}
        \includegraphics[width=\linewidth]{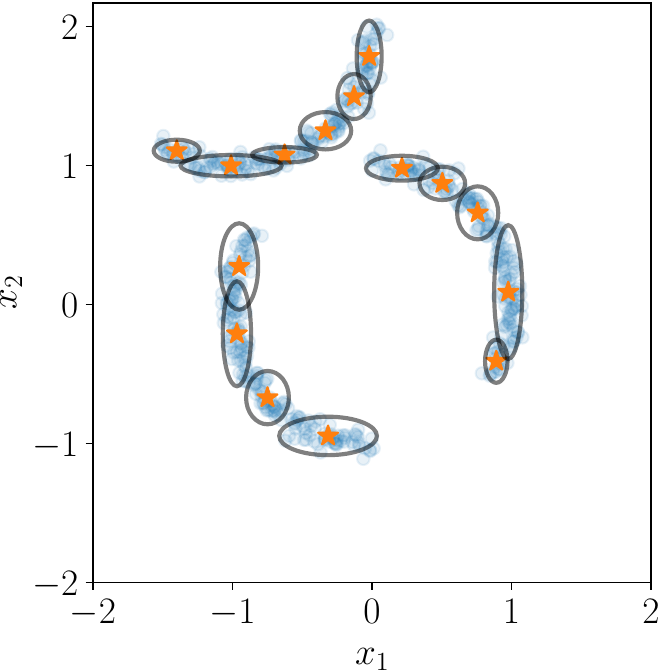}
        \caption{Offline fit.}
    \end{subfigure}\hfill
    \begin{subfigure}{0.32\linewidth}
        \includegraphics[width=\linewidth]{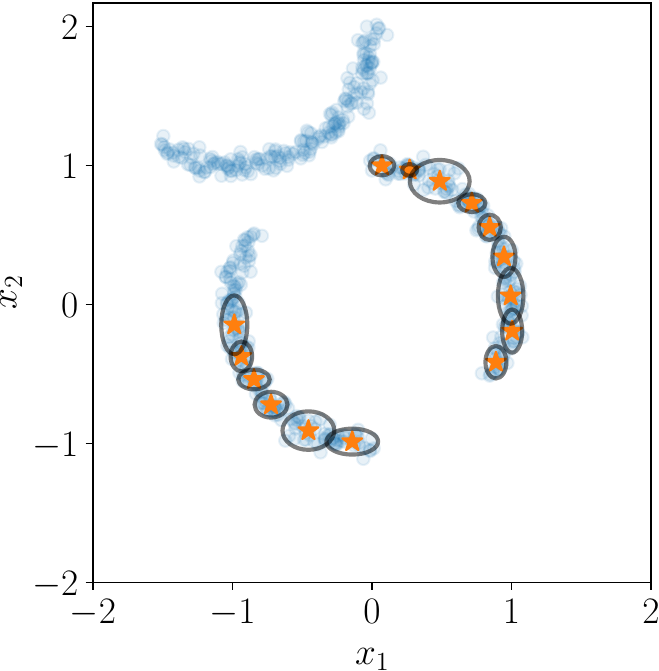}
        \caption{$it = 10$ ($\mathcal{W}_{2}$).}
    \end{subfigure}\hfill
    \begin{subfigure}{0.32\linewidth}
        \includegraphics[width=\linewidth]{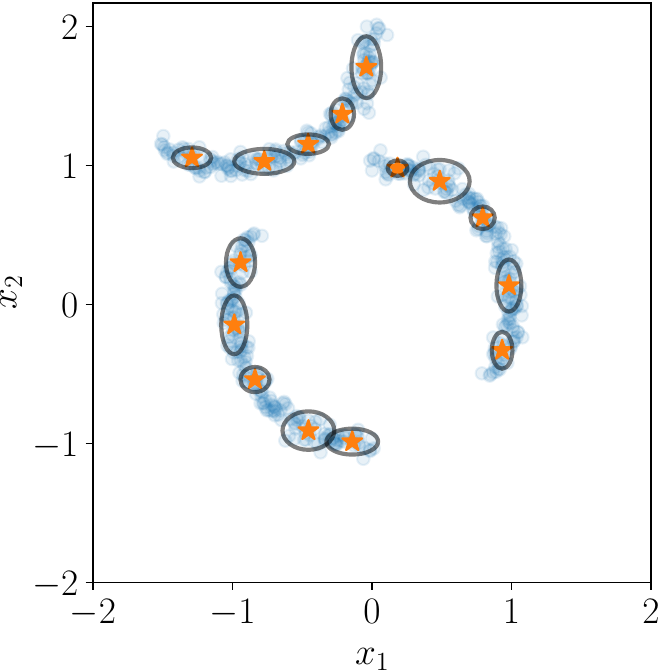}
        \caption{$it = 18$ ($\mathcal{W}_{2}$).}
    \end{subfigure}\\
    \caption{Toy example illustrating the online learning of a \gls{gmm} under the Kullback-Leibler divergence~\cite{acevedo2017multivariate} (b, c), and the $\mathcal{W}_{2}$ distance (ours, e, f). Overall, using our strategy we achieve a fit an offline \gls{gmm} (d).}
    \label{fig:ogmm_toy_example}
\end{figure}

In this section, we illustrate the advantage of our proposed online \gls{gmm} strategy in comparison with the method of~\cite{acevedo2017multivariate}. This toy example is shown in Fig.~\ref{fig:ogmm_toy_example} (a), which is composed of $3$ non-linear clusters, with $200$ samples each, amounting to a total of $n = 600$ samples. We encode the order in which samples arrive using a colormap. In this example, we set $K_{min} = 5$, $\Delta K = 3$ and $K_{max} = 15$. We use a batch size o $32$ samples, amounting to $19$ iterations.

We show the \gls{gmm} generated by~\cite{acevedo2017multivariate} as a function of the iteration $it$. As we show in Fig.~\ref{fig:ogmm_toy_example} (b, e), The two algorithms, ours and~\cite{acevedo2017multivariate}, initialize the \gls{gmm} in the same way. However, as we comment in section~\ref{sec:ogmm}, the main difference is that our algorithm is capable of updating Gaussian components. As we show in Fig.~\ref{fig:ogmm_toy_example} (c, f), this means that we can better accommodate components on new batches. As a result, as shown in Fig.~\ref{fig:ogmm_toy_example} (c, f), this leads to a \gls{gmm} that better resembles the offline fit, as can be seen Fig.~\ref{fig:ogmm_toy_example} (d).

\subsection{Online Multi-Source Domain Adaptation}

In this section, we experiment with online, multi-source domain adaptation, as explained in the beginning of section~\ref{sec:methodology}. Our goal is to learn a classifier on target domain, with access to offline source domain data, and target domain data that arrives in a stream of batches of $n_{b}$ samples. We experiment with the \gls{tep} benchmark~\cite{reinartz2021extended}, which consists of simulations of a complex, large scale chemical plant. In our experiments we use the setting of~\cite{montesuma2023multi}, that is, we first learn a convolutional neural net with source domain data, and use its encoder to extract a 128 dimensional feature vector. We refer readers to~\cite{montesuma2023multi} for further details.

The \gls{tep} benchmark includes $6$ domains, corresponding to different modes of operation. The statistical characteristics of the sensor readings change with the different modes. Meanwhile, there are $29$ classes, that is, $28$ faults and a normal state. We want to determine, from these sensor readings, which fault, or its absence, has occurred. Due to space constraints, we focus on the adaptation towards mode $1$. We divide target domain data in 5 independent partitions for performing $5-$fold cross-validation.

\begin{figure}[ht]
    \centering
    \begin{subfigure}{0.45\linewidth}
        \includegraphics[width=\linewidth]{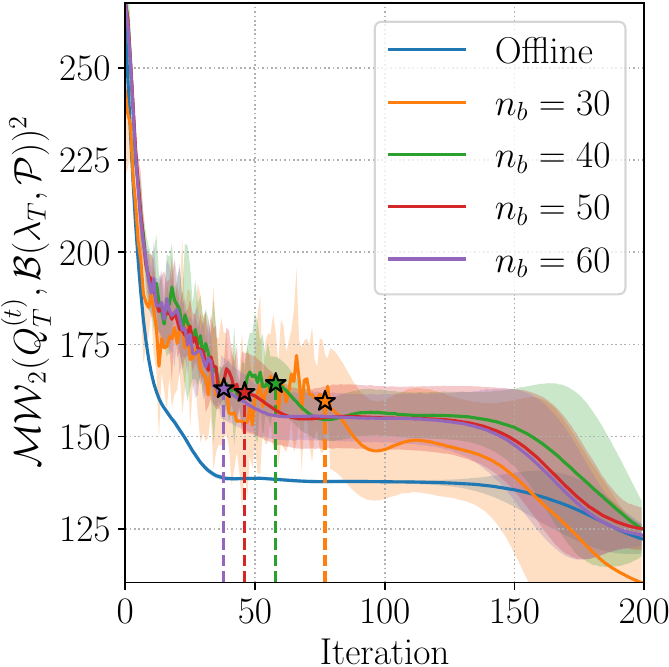}
        \caption{Reconstruction loss.}
    \end{subfigure}\hfill
    \begin{subfigure}{0.45\linewidth}
        \includegraphics[width=\linewidth]{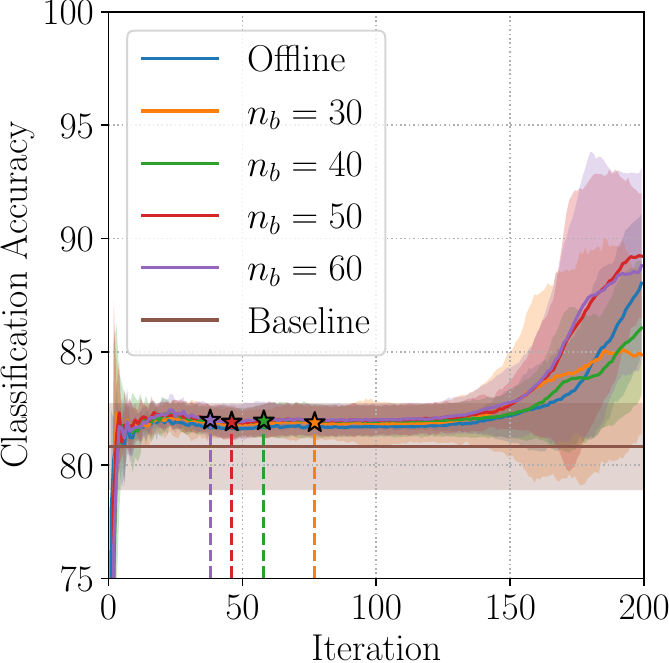}
        \caption{Classification accuracy.}
    \end{subfigure}
    \caption{Reconstruction loss and classification accuracy of \gls{ogmmdadil} on mode 1 (target domain). Experiments are run independently on 5 folds of the target domain data. Solid lines represent the average, and the shaded regions represent $\pm 2\sigma$ around the mean.}
    \label{fig:experiments_tep}
\end{figure}

We experiment with the online learning of a \gls{gmm} dictionary. As follows, we track the $\mathcal{MW}_{2}$ between $Q_{T}^{(t)}$ and its reconstruction $\mathcal{B}(\lambda_{T},\mathcal{P})$, shown in Fig.~\ref{fig:experiments_tep} (a). This quantity allows us to quantify how well we express the target \gls{gmm} as a mixture-Wasserstein barycenter of atom measures. Especially, we compare it with the offline version of \gls{gmm}-\gls{dadil}. Note that, in the initial iterations, learning is noisy due the updates in the target \gls{gmm}. We mark the end of the data stream with a colored star. Since we keep all \glspl{gmm} in memory, after the end of the updates in $Q_{T}^{(t)}$, we can continue optimization, which converge towards a local minimum similar to the offline algorithm.

We further compare \gls{gmm}-\gls{dadil} and our online version with the baseline, i.e., the performance of a classifier fit only with source domain data. In the case of \gls{gmm}-\gls{dadil}, we classify examples using the \gls{map} procedure explained in equation~(\ref{eq:map}). Our results are shown in Fig.~\ref{fig:experiments_tep} (b). With respect reconstruction loss, classification accuracy is more stable, with little fluctuation before the end of the data stream. As we see in the aforementioned figure, performance generally improves after the end of the data stream, showcasing the advantage of our online \gls{gmm} modeling.

\section{Conclusion}\label{sec:conclusion}

In this paper, we introduced novel methods towards online multi-source domain adaptation with a focus on cross-domain adaptation problems. In this case, we adapt heterogeneous historical datasets towards a stream of target domain data. Our main contributions are an online \gls{gmm} learning algorithm, and an online \gls{gmm}-\gls{dadil}~\cite{montesuma2024faster} algorithm. We experiment with the challenging \gls{tep} benchmark~\cite{reinartz2021extended,montesuma2023multi}, which contains simulations of a large-scale, complex chemical plant~\cite{reinartz2021extended,montesuma2023multi}. Our experimental results show that, through our methods, we can succesfully adapt \emph{on the fly} to target domain data. Furthermore, our online \gls{gmm} can serve as a memory representing the stream of target domain data, allowing dictionary learning to continue improving after the data stream ends. In future works, we consider performing class and task-incremental cross-domain fault diagnosis in a multi-source scenario.


\bibliographystyle{IEEEbib}
\bibliography{refs}

\end{document}

%% file: glossaries.tex
\newacronym{gmm}{GMM}{Gaussian Mixture Model}
\newacronym{msda}{MSDA}{Multi-Source Domain Adaptation}
\newacronym{dadil}{DaDiL}{Dataset Dictionary Learning}
\newacronym{em}{EM}{Expectation-Maximization}
\newacronym{bic}{BIC}{Bayesian Information Criterion}
\newacronym{ot}{OT}{Optimal Transport}
\newacronym{bcr}{BCR}{Barycentric Coordinates Regression}
\newacronym{obcr}{O-BCR}{Online BCR}
\newacronym{ogmm}{OGMM}{Online GMM}
\newacronym{odadil}{O-DaDiL}{Online DaDiL}
\newacronym{ogmmdadil}{O-GMM-DaDiL}{Online GMM DaDiL}
\newacronym{erm}{ERM}{Empirical Risk Minimization}
\newacronym{da}{DA}{Domain Adaptation}
\newacronym{tep}{TEP}{Tennessee Eastman Process}
\newacronym{map}{MAP}{Maximum a Posteriori}

%% file: algorithms/online_gmm_fit.tex
\begin{algorithm}[ht]
\caption{Online GMM fit.}
\label{alg:online_gmm_fit}
\Function{online\_gmm\_fit($stream, K_{min}, K_{max}, \Delta K$)}{
$P_{0} \leftarrow$ \texttt{\bfseries get\_best\_gmm($\mathbf{X}^{(P)}_{0}, K_{min}, K_{min}$)}\;
\For{$\mathbf{X}_{t}^{(P)}$ $in$ $stream$}{
$P \leftarrow$ \texttt{\bfseries get\_best\_gmm($\mathbf{X}^{(P)}_{t},1,\Delta K$)}\;
$P_{t} \leftarrow $ \texttt{\bfseries concat\_components($P_{t-1}, P$)}\;
$P_{t} \leftarrow$ \texttt{\bfseries compress\_gmm($P_{t}$)}\;
}
}
\Return{$P_{t}$}
\end{algorithm}

%% file: algorithms/get_best_gmm.tex
\begin{algorithm}[ht]
\caption{Fitting GMM to batch.}
\label{alg:get_best_gmm}
\Function{get\_best\_gmm($\mathbf{X}^{(P)}, k_{1}, k_{2}$)}{
$bic_{min} \leftarrow +\infty$\;
\For{$k=k_{1},\cdots,k_{2}$}{
    $P_{k} \leftarrow \text{EM}(\mathbf{X}^{(P)},k)$\;
    \If{$\text{BIC}(P) < bic_{min}$}{
        $P^{\star} \leftarrow P_{k}$\;
    }
}
}
\Return{$P^{\star}$}\;
\end{algorithm}

%% file: algorithms/compress_gmm.tex
\begin{algorithm}[ht]
\caption{Compressing GMM.}
\label{alg:compress_gmm}
\Function{compress\_gmm($P, K_{max}$)}{
\While{$|P| > K_{max}$}{
\For{$i=1,\cdots,|P|$}{
    $W_{ii} \leftarrow +\infty$\;
    \For{$j=i+1,\cdots,|P|$}{
        $W_{ij}, W_{ji} \leftarrow \mathcal{W}_{2}(P_{i},P_{j})$\;
    }
}
$i^{\star}, j^{\star} \leftarrow \text{argmin}_{i,j} W_{ij}$\;
$P_{i^{\star}} \leftarrow$ \texttt{\bfseries gauss\_merge($P, i^{\star}, j^{\star}$)}\;
\texttt{\bfseries delete($P, j^{\star}$)}\;
}
}
\Return{$P$}\;
\end{algorithm}

%% file: algorithms/merge_gaussians.tex
\begin{algorithm}[ht]
\caption{Combining Gaussian components.}
\label{alg:merge_gaussians}
\Function{gauss\_merge($P, i, j$)}{
$\lambda_{1} \leftarrow \frac{\pi_{i}^{(P)}}{\pi_{i}^{(P)}+\pi_{j}^{(P)}}$, $\lambda_{2} \leftarrow \frac{\pi_{j}^{(P)}}{\pi_{i}^{(P)}+\pi_{j}^{(P)}}$\;
$\pi \leftarrow \pi_{i}^{(P)} + \pi_{j}^{(P)}$\;
$\mu \leftarrow \lambda_1 \mu_{i}^{(P)} + \lambda_{2} \mu_{j}^{(P)}$\;
$\sigma \leftarrow \lambda_{1} \sigma_{i}^{(P)} + \lambda_{2} \sigma_{j}^{(P)}$\;
}
\Return{$\pi, \mu, \sigma$}\;
\end{algorithm}